\documentclass[10pt,twocolumn,letterpaper]{article}

\usepackage{iccv}
\usepackage{times}
\usepackage{epsfig}
\usepackage{graphicx}
\usepackage{amsmath}
\usepackage{amssymb}
\newtheorem{theorem}{Theorem}
\usepackage{subfig}

\usepackage{algorithm}
\usepackage{algpseudocode}
\usepackage{multirow}
\usepackage[font=normalsize,skip=5pt]{caption}
\usepackage[accsupp]{axessibility}

\usepackage[pagebackref=true,breaklinks=true,colorlinks,bookmarks=false]{hyperref}

\iccvfinalcopy 

\def\iccvPaperID{8813}

\ificcvfinal\fi

\begin{document}

\title{Score Priors Guided Deep Variational Inference  \\ for Unsupervised Real-World Single Image Denoising}

\author{Jun Cheng, Tao Liu, Shan Tan\footnotemark[1]\\
Huazhong University of Science and Technology, Wuhan, China\\
{\tt\small \{jcheng24, hust\_liutao, shantan\}@hust.edu.cn}
}

\maketitle

\ificcvfinal\thispagestyle{empty}\fi

\renewcommand{\thefootnote}{\fnsymbol{footnote}}
\footnotetext[1]{Corresponding author.}

\begin{abstract}
Real-world single image denoising is crucial and practical in computer vision. Bayesian inversions combined with score priors now have proven effective for single image denoising but are limited to white Gaussian noise. Moreover, applying existing score-based methods for real-world denoising requires not only the explicit train of score priors on the target domain but also the careful design of sampling procedures for posterior inference, which is complicated and impractical. To address these limitations, we propose a score priors-guided deep variational inference, namely ScoreDVI, for practical real-world denoising. By considering the deep variational image posterior with a Gaussian form, score priors are extracted based on easily accessible minimum MSE Non-$i.i.d$ Gaussian denoisers and variational samples, which in turn facilitate optimizing the variational image posterior. Such a procedure adaptively applies cheap score priors to denoising. Additionally, we exploit a Non-$i.i.d$ Gaussian mixture model and variational noise posterior to model the real-world noise. This scheme also enables the pixel-wise fusion of multiple image priors and variational image posteriors. Besides, we develop a noise-aware prior assignment strategy that dynamically adjusts the weight of image priors in the optimization. Our method outperforms other single image-based real-world denoising methods and achieves comparable performance to dataset-based unsupervised methods.
\end{abstract}

\section{Introduction}
Image denoising is a fundamental task in low-level computer vision. Different from the additive white Gaussian noise (AWGN) that has been widely studied, real-world noise is typically signal-dependent and spatially correlated (or structured) and is prevalent in various image processing applications, including photography with sRGB noise \cite{cFlow, APBSN} and bioimaging with microscopy image noise \cite{HDN}. The removal of real-world structured noise is therefore critical for the subsequent image analysis and understanding.  

While recent deep learning-based supervised methods \cite{DNCNN, SwinIR, Restormer, NAFNet} and unsupervised/self-supervised methods \cite{CVF-SID,APBSN,LUD_VAE,cFlow} have shown promising results on real-world image denoising, they suffer from data collection difficulties and model generalization issues \cite{zheng2020unsupervised}. As an alternative, single image-based real-world denoising methods that rely on only a single image are gaining increasing attention as a more practical and promising solution. However, existing single image-based methods, e.g., \cite{DIP, Noise2self, Self2self, PD, structn2v, zheng2020unsupervised, noise2fast}, show unsatisfactory performance in real-world structured noise removal, which highlights the need for new and effective approaches. 

Nowadays, deep generative models, represented by score-based methods \cite{score_sde, DDPM}, have enabled powerful image priors to be learned from large-scale clean datasets. The learned
score priors from score networks have demonstrated remarkable capacity in image generation and various image restorations \cite{DDPM_DDRM,score_sde,score_CT,stable_diffusion}. However, current score-based methods related to denoising, e.g., \cite{score_CSMRI,score_imagedenoise,DDPM_DDRM,score_SNIPS,score_pet,score_MRI_3}, typically consider AWGN and cannot handle real-world noise. Furthermore, applying current score-based methods for real-world denoising not only requires the explicit train of score priors on the target domain but also has to devise sampling procedures elaborately for posterior inference. The accurate noise distribution is also demanded in the conditional sampling. Above restrictions makes them cumbersome and impractical for real-world noise removal. 

To address these limitations, we propose an effective and practical score priors guided deep variational inference, namely ScoreDVI, for unsupervised real-world single image denoising. Rather than explicitly training score priors on the target domain, we relate the general score functions to the easily available minimum MSE (MMSE) Non-$i.i.d$ Gaussian denoisers. We extract score priors by employing well-trained Non-$i.i.d$ Gaussian denoisers to denoise samples from the variational image posterior paramterized by deep neural networks and then use the extracted scores to 
facilitate optimizing this variational image posterior, which in turn allows adaptive extraction of scores in the next round. This optimization procedure adaptively applies cheap score priors to denoising and frees the elaborate design of sampling strategies. 

Instead of relying on the accurate noise model, we adopt the Non-\textit{i.i.d} Gaussian mixture model (GMM) to roughly represent real-world noise, and a variational noise posterior is  built and optimized to improve the modeling accuracy by means of hieratical Bayesian inference. This modeling strategy also allows for the pixel-wise fusion of multiple image priors and variational image posteriors, further enhancing the denoising capability. Moreover, traditional variational inference treats image priors and likelihoods equally during optimization, which may not be optimal for inputs with varying levels of noise. Since noisier images require more priors, we propose a noise-aware prior assignment strategy that dynamically assigns different weights of image priors during optimization based on the estimated noise levels of noisy images. By doing so, our method can maximize performance in individual instances.  

In total, our contributions are summarized as follows:
\begin{enumerate}
    \item We propose an adaptive extraction and utilization of score priors based on Non-\textit{i.i.d} Gaussian denoisers and deep variational inference for variational optimization.
    \item We incoperate Non-\textit{i.i.d} GMM likelihood model and the variational noise posterior to model the real-world noise, which enables pixel-wise fusion of multiple image priors and variational image posteriors.
    \item We develop a noise-aware prior assignment strategy, which adaptively assigns the degree of image priors to the optimization objective for different noisy images.
    \item To the best of our knowledge, we are the first to successfully remove real-world structured noise by applying score priors. Our method outperforms other single image-based methods on a variety of benchmark datasets and achieves competitive results against dataset-based unsupervised methods.
\end{enumerate}

\section{Related works}
\subsection{Real-world image denoising}
\noindent \textbf{Dataset-based unsupervised learning methods}. To address the data collection difficulties of supervised approaches \cite{MPRNet, SwinIR, Restormer, NAFNet}, several unsupervised and self-supervised real-world denoising methods have been proposed. Blind-spot (BS) based strategies, e.g., Noise2self \cite{Noise2self}, Nei2nei \cite{Neighbor2neighbor}, and BS networks \cite{SSID, DBSN} as well as their zero-shot versions, are limited to handling pixel-independent noise. Noiser2noise \cite{Noisier2noise}, R2R \cite{R2R}, and IDR \cite{IDR} can address structured noise, but they require accurate noise distribution, making them impractical. 

Recently, \cite{PD} proposed using pixel-shuffle downsampling (PD) to break down the spatial correlation of structured noise. AP-BSN \cite{APBSN} used two PDs with different strides for training and testing. CVF-SID \cite{CVF-SID} attempted to separate the latent image and noise from the noisy input by a cyclic module and self-supervised losses. LUD-VAE \cite{LUD_VAE} designed two hidden variables for the latent image domain and noise domain, respectively, and derived an optimizable loss for unpaired data. \cite{cFlow, C2N, autoregressive_noise_model} explicitly learned the real-world noise distribution. These methods require noisy or unpaired data and generalization problems remain. 

\noindent \textbf{Single image-based real-world denoising}. 
Deep image prior (DIP) \cite{DIP} achieved single image denoising by regularizing the solution via CNN architectures. Self2Self \cite{Self2self} built training pairs from a single noisy image via Bernoulli sampling to train denoisers. \cite{PD} proposed a pixel-shuffle downsampling (PD) to break down the spatial correlation of real-world structured noise. StructN2V \cite{structn2v} estimated the noise correlation from a single image and developed a structured mask to apply N2V \cite{Noise2void}. NN+denoiser \cite{zheng2020unsupervised} used conditional VAE to model the likelihood of real-world noise and combined plug-and-play priors to denoise. However, these methods show poor performance when removing severe structured noise. Instead, our ScoreDVI can denoise noisier images effectively and efficiently.   

\subsection{Image priors modeling}
Classical image priors \cite{BIV_TV,Nonlocal,sparsity_prior,lowrank_prior} utilize low-level statistics of natural images to denoise images. Recently, complex image priors can be learned by using deep generative models and are then employed in Bayesian inversion to solve various image restoration tasks \cite{VAE_prior, GAN_prior, DDPM_DDRM}. Particularly, score-based methods have achieved remarkable performance \cite{DDPM_DDRM,score_sde,stable_diffusion}. However, current research on image restoration with score priors \cite{score_imagedenoise,score_SNIPS,score_implicit_prior,score_CT,score_MRI,score_mri_2,DDPM_DDRM,score_pet,score_MRI_3} generally consider AWGN or noiseless inputs. And the direct application of these methods for real-world noise is complicated and impractical. Recently, \cite{score_structured_noise} used the diffusion model to remove structured noise, while their method simply extended the existing paradigm and can only handle simple cases such as removing digits from face images.  Instead, we propose to combine score priors with MMSE Gaussian denoisers and variational inference, which allows us to effectively and efficiently remove real-world noise.

\subsection{Variational inference}
Traditional variational inference-based methods \cite{BIV_VI1, BIV_VI2, BIV_VI3} employ analytical image priors and cyclic optimization to obtain the approximate posterior, but they often suffer from poor performance. Recently, deep learning-based variational inference methods for real-world image denoising such as VDN \cite{VDN}, VNIR \cite{VNIR}, and VDIR \cite{VDIR} construct variational posteriors via DNNs and utilize ground truths as strong priors. However, these methods require paired training data. Instead, our ScoreDVI is unsupervised and can deal with structured noise by employing score priors.

\section{Method}
In this section, we present our unsupervised deep variational inference method for real-world single image denoising. We first introduce the variational inference framework in section \ref{Variational_inference}, followed by the idea of adaptively applying score priors to facilitate the optimization of variational objective in section \ref{Deep_score_priors_for_sRGB_noisy_images}. The Non-$i.i.d$ Gaussian mixture likelihood model and hyperpriors are presented in section \ref{likelihood_model}. Section \ref{Deep_modeling_of_variational_posteriors} shows the deep variational posteriors and the complete optimization objective. The noise-aware prior assignment is finally given in section \ref{Noise-aware_prior_assignment}.

\subsection{Variational inference framework}\label{Variational_inference}

In Bayesian statistics, image denoising is viewed as finding the posterior distribution $p(\boldsymbol{x}|\boldsymbol{y})$ and its estimators, e.g., posterior mean and MAP solution, where $\boldsymbol{x},\boldsymbol{y}\in \mathbb{R}^N$ are latent clean image and noisy observation, respectively. $N=C\times H\times W$ is the image dimension. 
In practice, the posterior distribution $p(\boldsymbol{x}|\boldsymbol{y})$ is non-trivial and has no closed-form solution. The idea of variational inference is to  construct a tractable distribution $q(\boldsymbol{x})$ and minimize its `distance` to the true $p(\boldsymbol{x}|\boldsymbol{y})$.  KL divergence is usually employed as the distance metric, 
\begin{equation}\label{ELBO_simple}
\begin{aligned}
    &\text{KL}(q(\boldsymbol{x}) || p(\boldsymbol{x}|\boldsymbol{y})) = \int \log \frac{q(\boldsymbol{x})}{p(\boldsymbol{x}|\boldsymbol{y})}q(\boldsymbol{x})d\boldsymbol{x} \\=&
    \log p(\boldsymbol{y}) - \underbrace{\left( \text{E}_{q(\boldsymbol{x})}  (\log p(\boldsymbol{y}|\boldsymbol{x})) - \text{KL}(q(\boldsymbol{x}) || p(\boldsymbol{x})) \right)}_{\text{ELBO}}
\end{aligned}
\end{equation}
where $\log p(\boldsymbol{y})$ is the marginal likelihood (also called evidence), and $\text{E}_{q(\boldsymbol{x})}  (\log p(\boldsymbol{y}|\boldsymbol{x})) -\text{KL}(q(\boldsymbol{x}) || p(\boldsymbol{x}))$ is the Evidence Low Bound (ELBO) as $\text{KL}(q(\boldsymbol{x}) || p(\boldsymbol{x}|\boldsymbol{y})) \geq 0$. 

As $\log p(\boldsymbol{y})$ is generally not computable, the ELBO instead acts as the objective of variational inference. Hence, modeling the image prior $p(\boldsymbol{x})$, likelihood  $p(\boldsymbol{y}|\boldsymbol{x})$, and variational posterior $q(\boldsymbol{x})$ involved in the ELBO is important.

\subsection{Adaptive score priors for variational inference}\label{Deep_score_priors_for_sRGB_noisy_images} 
In this section, we propose an adaptive extraction and application of score priors in variational inference. 

\noindent \textbf{Scores and MMSE Gaussian denoisers}.
\cite{Score_GaussianDenoiser} derived the relationship between scores and MMSE denoisers for $i.i.d$ Gaussian noise.
Here, we extend such a relationship to the case of general structured Gaussian. 

\begin{theorem}\label{score_proof}
Suppose a noisy observation $\widetilde{\boldsymbol{x}}=\boldsymbol{x}+\boldsymbol{n} \in \mathbb{R}^{N}$ where $\boldsymbol{n}\sim \mathcal{N}(\boldsymbol{0}, \Sigma)$ has the zero mean and covariance $\Sigma$, and $\boldsymbol{x} \sim p(\boldsymbol{x})$. Then the score of $p(\widetilde{\boldsymbol{x}})$ is 
\begin{equation} \label{multi-score}
    \nabla_{\widetilde{\boldsymbol{x}}} \log p(\widetilde{\boldsymbol{x}}) = \Sigma^{-1} \left(\text{E}_{p({\boldsymbol{x}}|\widetilde{\boldsymbol{x}})}(\boldsymbol{x}) - \widetilde{\boldsymbol{x}}\right)
\end{equation}
where $\text{E}_{p({\boldsymbol{x}}|\widetilde{\boldsymbol{x}})}(\boldsymbol{x})$ is the conditional mean for input $\widetilde{\boldsymbol{x}}$ corrupted by general Gaussian noise.
\end{theorem}

The proof of Theorem \ref{score_proof} is given in Supplementary Material S1.1. If we further consider the noise covariance has diagonal form $\Sigma = \text{diag}(\boldsymbol{\sigma}^2)$, Eq. \eqref{multi-score} becomes 
\begin{equation} \label{score_diag}
\nabla_{\widetilde{\boldsymbol{x}}} \log p(\widetilde{\boldsymbol{x}}) = \frac{1}{\boldsymbol{\sigma}^2} \odot \left(\text{E}_{p({\boldsymbol{x}}|\widetilde{\boldsymbol{x}})}(\boldsymbol{x}) - \widetilde{\boldsymbol{x}}\right) 
\end{equation}
where $\odot$ denotes element-wise multiplication.

In practice, the posterior mean $\text{E}_{p({\boldsymbol{x}}|\widetilde{\boldsymbol{x}})}(\boldsymbol{x})$ in Eq. \eqref{score_diag} can be approximated by the output of a deep  MMSE Non-$i.i.d$ (independent but not indentical) Gaussian denoiser $\mathcal{G}(\widetilde{\boldsymbol{x}})$. Moreover, a CNN-based blind MMSE Gaussian denoiser trained over AWGN with a wide range of noise variance has the ability to handle such Non-$i.i.d$ noise as the local connectivity property of CNNs. Therefore, score priors are directly available from well-trained blind MMSE Gaussian denoisers, freeing the explicit training of score networks. 

\noindent \textbf{Scores facilitate ELBO optimization}. Within the ELBO objective in Eq. \eqref{ELBO_simple}, the image prior $p(\boldsymbol{x})$ is only involved in the KL divergence term of the ELBO, which can be approximated as 
\begin{equation}\label{KL_MC}
\begin{aligned}
    \text{KL}(q(\boldsymbol{x}) & || p(\boldsymbol{x})) = \text{E}_{q(\boldsymbol{x})} \left( \log q(\boldsymbol{x}) - \log p(\boldsymbol{x})\right)  \\ &\approx \text{E}_{q(\boldsymbol{x})}\log q(\boldsymbol{x}) - \frac{1}{M}\sum_{m=1}^M \log p(\boldsymbol{x}_m)
\end{aligned}
\end{equation}
where the Mente Carlo (MC) integration with $M$ samples is used to approximate $\text{E}_{q(\boldsymbol{x})}\log p(\boldsymbol{x})$, and $\boldsymbol{x}_m \sim q(\boldsymbol{x})$. 

If we take the derivative of $\text{KL}(q(\boldsymbol{x}) || p(\boldsymbol{x}))$ in Eq. \eqref{KL_MC} with regard to $\boldsymbol{x}$, then the score function appears. In order to incorporate the score prior from Gaussian denoisers, we consider that $q(\boldsymbol{x})$ follows diagonal Gaussian distribution $\mathcal{N}(\boldsymbol{\mu}, \text{diag}(\boldsymbol{\sigma}^2))$, 
and then the derivatives of $\text{E}_{q(\boldsymbol{x})}  \log p(\boldsymbol{x})$ with regard to variational posterior parameters $\boldsymbol{\mu}$ and $\boldsymbol{\sigma}^2$ can be computed as
{\fontsize{9.5pt}{11.5pt}\selectfont
\begin{equation}\label{score_KL}
\begin{aligned}
    &\nabla_{\boldsymbol{\mu}} \text{E}_{q(\boldsymbol{x})}  \log p(\boldsymbol{x})  \approx \frac{1}{M}\sum_{m=1}^M({\mathcal{G}(\boldsymbol{x}_m)-\boldsymbol{x}_m}) \odot \frac{1}{\boldsymbol{\sigma}^2} \\
    &\nabla_{\boldsymbol{\sigma}^2}\text{E}_{q(\boldsymbol{x})}  \log p(\boldsymbol{x}) \approx \frac{1}{M}\sum_{m=1}^M({\mathcal{G}(\boldsymbol{x}_m)-\boldsymbol{x}_m}) \odot \frac{1}{\boldsymbol{\sigma}^2} \odot \boldsymbol{\epsilon} \\
    &\boldsymbol{x}_m = \boldsymbol{\mu} + \boldsymbol{\sigma} \odot \boldsymbol{\epsilon}, \boldsymbol{\epsilon} \sim \mathcal{N}(\boldsymbol{0}, I)
\end{aligned}
\end{equation}}

The derivation of Eq. \eqref{score_KL} is given in Supplementary Material S1.2. Eq. \eqref{score_KL} reveals that there is no need to compute $\text{E}_{q(\boldsymbol{x})}  \log p(\boldsymbol{x})$ explicitly, instead, we can utilize the score prior from  MMSE Non-$i.i.d$ Gaussian denoisers to perform the gradient-based optimization of $\text{KL}  (q(\boldsymbol{x}) || p(\boldsymbol{x}))$ in the ELBO objective. Moreover, the extraction of scores from MMSE Gaussian denoisers is adaptive, and we do not need to explicitly schedule the noise variance and design the sampling strategy as done in score-based generative models \cite{score_sde,score_SNIPS,DDPM_DDRM}. That is, the score for
$\boldsymbol{x}_m(t) \sim \mathcal{N}(\boldsymbol{\mu}(t), \boldsymbol{\sigma}^2(t))$ in the $t$-th iteration assists in the update of $\boldsymbol{\mu}(t+1)$ and $ {\boldsymbol{\sigma}^2(t+1)}$ , which in turn helps extract the score for $\boldsymbol{x}_m(t+1)$ dynamically in the $(t+1)$-th iteration.  

\subsection{Non-\textit{i.i.d} Gaussian mixture likelihood model}\label{likelihood_model}
The real-world structured noise is generally signal-dependent and spatially correlated, which differs greatly from AWGN. In order to incorporate our knowledge about real-world noise, we utilize the Non-\textit{i.i.d}  Gaussian mixture model (GMM) to model the likelihood function, which is
{\fontsize{9.5pt}{11.4pt}\selectfont
\begin{equation}\label{GMM1}
p(\boldsymbol{y}|X,\Phi,\Omega) = \prod_{i=1}^N \sum_{k=1}^K \omega_{ik} \mathcal{N}(x_{ik}, {\phi}_{ik}^{-1}), \sum_{k=1}^K \omega_{ik}=1
\end{equation}}
where $X=[\boldsymbol{x}_1,\cdots\boldsymbol{x}_K ] ,\Phi=[\boldsymbol{\phi}_1,\cdots\boldsymbol{\phi}_K ] , \Omega=[\boldsymbol{\omega}_1,\cdots\boldsymbol{\omega}_K ] \in \mathbb{R}^{N \times K}$ are matrices with $K$ columns, respectively. ${x}_{ik}$ in $X$, ${\phi}_{ik}$ in $\Phi$, and ${\omega}_{ik}$ in $\Omega$ denote the mean, precision, and mixing parameter of the \textit{k}-th Gaussian component, respectively, for modeling $i$-th pixel $y_i$ of $\boldsymbol{y}$. $K$ is the total number of mixtures.

In Eq. \eqref{GMM1}, each noisy pixel $y_i$ is represented by a GMM, so as to better model the spatially variant characteristic of real-world noise. Besides, such consideration 
allows the ensemble of different  image priors and variational image posteriors in a pixel-wise manner, which will be introduced in the following section. To facilitate optimization, we refer to \cite{noniidmodeling} and reparameterize the likelihood function in Eq. \eqref{GMM1} as
\begin{equation}\label{GMM2}
\begin{aligned}
    p(\boldsymbol{y}|X, Z, \Phi) &= \prod_{i=1}^N \prod_{k=1}^K \mathcal{N}(x_{ik}, {\phi}_{ik}^{-1})^{z_{ik}}\\p(Z|\Omega)&=\prod_{i=1}^N \prod_{k=1}^K \omega_{ik}^{z_{ik}}, \sum_{k=1}^K \boldsymbol{z}_{ik}=1
\end{aligned}
\end{equation}
where $Z=[\boldsymbol{z}_1,\cdots\boldsymbol{z}_K ]\in \mathbb{R}^{N\times K}$ is an auxiliary matrix variable and each row of $Z$ follows a categorical distribution with one item one and the others zero. 

\noindent \textbf{Hyperpriors for hyperparameters}.
Different from existing score-based methods which carry out posterior sampling based on accurate likelihood functions with fixed distribution parameters, we view $\Phi$ and $\Omega$ in Eq. \eqref{GMM2} as random variables and infer them by introducing hyperpriors based on hieratical Bayesian inference. In order to simplify the ELBO objective, we choose the following conjugate priors for $\Phi$ and $\Omega$, respectively
\begin{equation}\label{hyperprior}
\begin{aligned}
   &p(\Phi|Z) = \prod_{i=1}^N \prod_{k=1}^K \text{Gamma}(\phi_{ik}; \alpha, \beta)^{z_{ik}}
   \\&
    p(\Omega)= \prod_{i=1}^N Z_{\text{Dir}}(\boldsymbol{d})\prod_{k=1}^K \omega_{ik}^{d_{k}-1}, \boldsymbol{d}\in \mathbb{R}^K
\end{aligned}
\end{equation}
where $\text{Gamma}$ denotes the Gamma distribution with parameters $\alpha$ and $\beta$. Each row of $\Omega$ follows the Dirichlet distribution with parameter $\boldsymbol{d}$ and normalization coefficient $Z_{\text{Dir}}(\cdot)$. 

\begin{figure*}[!htb]
    \centering
    \includegraphics[width=0.94\textwidth]{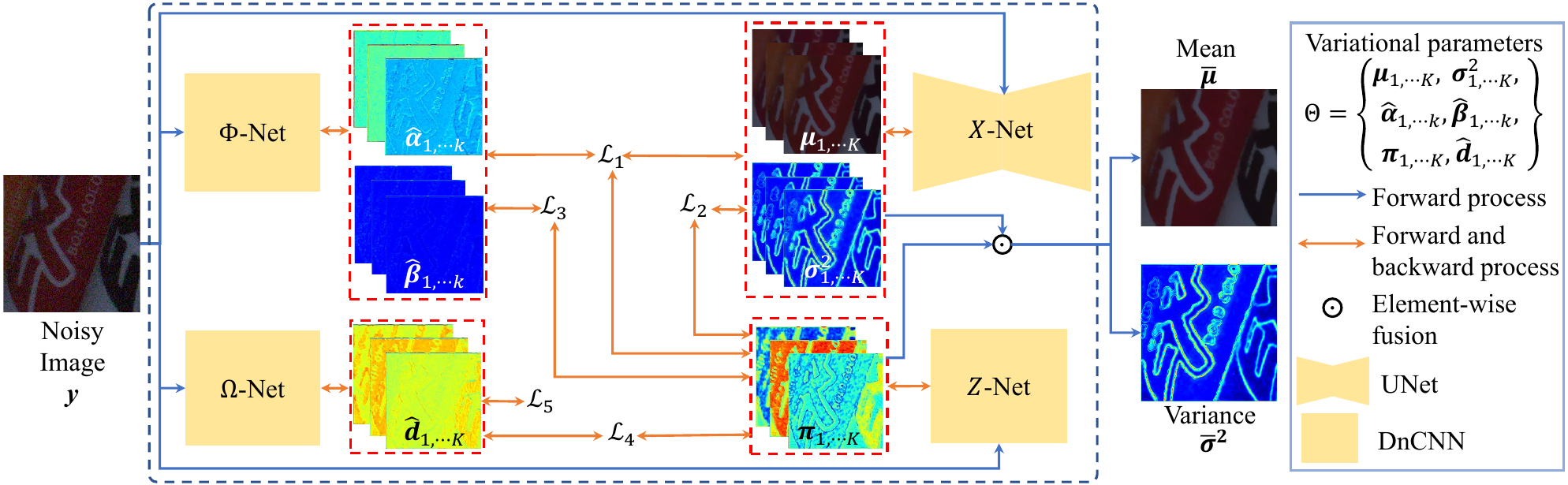}
    \caption{Overview of the proposed ScoreDVI. Our method receives a noisy image $\boldsymbol{y}$ and outputs the denoised image $\boldsymbol{\Bar{\mu}}$ as well as its variance $\Bar{\boldsymbol{\sigma}}^2$ when the optimization is done. The four networks are used to output the variational parameters $\Theta$. }
    \label{fig:main_diagram}
\end{figure*}

\subsection{Deep modeling of variational posteriors}\label{Deep_modeling_of_variational_posteriors}
The modeling of likelihood function in Eq. \eqref{GMM2} enables us to build the factorized image prior conditioned on $Z$ as 
\begin{equation}\label{prior_ensemble}
    p(X|Z) = \prod_{i=1}^N \prod_{k=1}^K p(x_{ik})^{z_{ik}}
\end{equation}
which enables the image prior $p(X)$ to be modeled by fusing many priors in a pixel-wise manner.

Considering Eqs. \eqref{GMM2}, \eqref{hyperprior} and \eqref{prior_ensemble}, the posterior distribution we want to infer now is $p(X,\Phi,Z,\Omega|\boldsymbol{y})$, while it is non-trivial. Therefore, we propose a powerful and trivial deep variational posterior $q(X,\Phi,Z,\Omega)$ that approximates the true $p(X,\Phi,Z,\Omega|\boldsymbol{y})$.

Specifically, we factorize the variational posterior $q(X,\Phi,Z,\Omega)$ as $q(X|Z)q(\Phi|Z)q(Z)q(\Omega)$ with each item
{\fontsize{9.5pt}{11.4pt}\selectfont
\begin{equation}\label{variational_posterior}
\begin{aligned}
    & q(X|Z) = \prod_{i=1}^N\prod_{k=1}^K \mathcal{N}(\mu_{ik}, \sigma^2_{ik})^{z_{ik}} 
    \\& q(\Phi|Z) = \prod_{i=1}^N\prod_{k=1}^K  \text{Gamma}(\phi_{ik}; \hat{\alpha}_{ik}, \hat{\beta}_{ik})^{z_{ik}}\\
    &q(Z)=\prod_{i=1}^N\prod_{k=1}^K \pi_{ik}^{z_{ik}},
    q(\Omega) = \prod_{i=1}^NZ_\text{Dir}(\boldsymbol{\hat{d}}_i)\prod_{k=1}^K \omega_{ik}^{\hat{d}_{ik}-1}
\end{aligned}
\end{equation}
}

All involved parameters of variational posteriors in Eq. \eqref{variational_posterior}, i.e., $\Theta = \{\mu_{ik},\sigma^2_{ik},\hat{\alpha}_{ik}, \hat{\beta}_{ik},\pi_{ik},{\hat{d}}_{ik}| i\in \{1,\cdots N \}, k\in \{1,\cdots K \} \}$ are modeled by CNNs as they provide good regularization for $\Theta$. Note that $q(\Phi)$ is variational noise posterior. As shown in Figure \ref{fig:main_diagram}, we build four CNNs, i.e., $X$-net, $\Phi$-net, $Z$-net and $\Omega$-net, which have $\boldsymbol{y}$ as the input and output variational posterior parameters $\Theta$.

Combing Eq. \eqref{GMM2}, Eq. \eqref{prior_ensemble}, Eq. \eqref{hyperprior} and Eq. \eqref{variational_posterior}, we can derive the complete optimization objective, which is
\begin{subequations}\label{ELBO}
\begin{align}
\mathcal{L}= &
-\text{ELBO} \\= &-\text{E}_{q(X, \Phi, Z)}(\log p(\boldsymbol{y}|X,\Phi,Z)) \rightarrow \mathcal{L}_1 \\& +\text{E}_{q(Z)}\left(\text{KL}(q(X|Z||p(X|Z)\right) \rightarrow \mathcal{L}_2 \\& + \text{E}_{q(Z)}\left(\text{KL}(q(\Phi|Z||p(\Phi|Z)\right) \rightarrow \mathcal{L}_3 \\&  +\text{E}_{q(\Omega)} \left(\text{KL}(q(Z||p(Z|\Omega)\right) \rightarrow \mathcal{L}_4\\& + \text{KL}(q(\Omega||p(\Omega))\rightarrow \mathcal{L}_5
\end{align}
\end{subequations}
with each component as follows:
\begin{equation}
\begin{aligned}
    \mathcal{L}_1 = -\sum_{i=1}^N\sum_{k=1}^K {\pi}_{ik} & \left ( -\frac{\left ( (\mu_{ik}-y_i)^2 + \sigma_{ik}^2 \right)\hat{\alpha}_{ik}}{2\hat{\beta}_{ik}} \right.  \\ & \left. + \frac{\psi(\hat{\alpha}_{ik})-\log \hat{\beta}_{ik}}{2} \right)
\end{aligned}
\end{equation}
\vspace{-6mm}
\begin{equation}
\begin{aligned}
\mathcal{L}_2  =  \sum_{i=1}^{N}\sum_{k=1}^{K} \left(
 -\frac{1}{2} \pi_{ik}  \log \sigma^2_{ik} - \pi_{ik}  \text{E}_{q(x_{ik})}  \log p(x_{ik}) \right)
\end{aligned}
\end{equation}
\vspace{-3mm}
\begin{equation}
\begin{aligned}
\mathcal{L}_3=
\sum_{i=1}^N\sum_{k=1}^K {\pi}_{ik} & \left ( \log
\frac{\Gamma(\alpha)}{\Gamma(\hat{\alpha}_{ik})} + 
(\hat{\alpha}_{ik} - \alpha) \psi(\hat{\alpha}_{ik}) \right.  \\ & \left. + \alpha\log \frac{\hat{\beta}_{ik}}{\beta} + (\beta - \hat{\beta}_{ik})\frac{\hat{\alpha}_{ik}}{\hat{\beta}_{ik}} \right)
\end{aligned}
\end{equation}
\vspace{-2mm}
\begin{equation}
\begin{aligned}
\mathcal{L}_4=  \sum_{i=1}^N\sum_{k=1}^K {\pi}_{ik} \left( \log {\pi}_{ik} - \psi(\hat{d}_{ik}) + \psi(\sum_{k=1}^K \hat{d}_{ik}) \right)
\end{aligned}
\end{equation}
\vspace{-2mm}
\begin{equation}
\begin{aligned}
\mathcal{L}_5 & =  \sum_{i=1}^N\sum_{k=1}^K (\hat{d}_{ik}-d_k)(  \psi(\hat{d}_{ik}) - \psi(\sum_{k=1}^K \hat{d}_{ik})) \\ & + \sum_{i=1}^N\log \frac{Z_\text{Dir}(\boldsymbol{\hat{d}}_{i})}{Z_\text{Dir}(\boldsymbol{d})}
\end{aligned}
\end{equation}
where  $\Gamma(\cdot)$ and $\psi(\cdot)$ denote gamma function and digamma function, respectively; For detailed derivations, please refer to Supplementary Material S1.3.

Note that $\mathcal{L}_1,\mathcal{L}_3,\mathcal{L}_4, \mathcal{L}_5$ and $ -\frac{1}{2} \pi_{ik}  \log \sigma^2_{ik}$ in $\mathcal{L}_2$ are  all directly computable and analytically derivable. Regarding $\text{E}_{q(x_{ik})}  \log p(x_{ik})$ in $\mathcal{L}_2$, due to the factorized $p(X|Z)$ and $q(X|Z)$, we have
\begin{equation}\label{score_KL2}
\begin{aligned}
\nabla_{\mu_{ik}}\text{E}_{q(x_{ik})}  \log p(x_{ik}) &= \left(\nabla_{\boldsymbol{\mu}_{k}}\text{E}_{q(\boldsymbol{x}_{k})} \log p(\boldsymbol{x}_{k}) \right)_i \\
\nabla_{\sigma^2_{ik}}\text{E}_{q(x_{ik})}  \log p(x_{ik}) &= \left (\nabla_{\boldsymbol{\sigma}^2_{k}}\text{E}_{q(\boldsymbol{x}_{k})}  \log p(\boldsymbol{x}_{k})\right)_i 
\end{aligned}
\end{equation} 
which can then be directly computed by using Eq. \eqref{score_KL}. 

The diagram and the whole pipeline of the proposed ScoreDVI are illustrated in Figure \ref{fig:main_diagram} and Algorithm \ref{alg}, respectively. Once the optimization of Eq. \eqref{ELBO} is done, we can obtain the mean and variance of $q(X)$ from $q(X|Z)$ and $q(Z)$:
\begin{equation}\label{posterior_ensemble}
    \Bar{\boldsymbol{\mu}}=
    \sum_{k=1}^K \boldsymbol{{\pi}}_k \odot \boldsymbol{\mu}_k, \Bar{\boldsymbol{\sigma}}^2=
    \sum_{k=1}^K \boldsymbol{{\pi}}^2_k \odot \boldsymbol{\sigma}^2_k 
\end{equation}
which turn out to be the pixel-wise fusion of different variational image posteriors $\boldsymbol{x}_k$. $\Bar{\boldsymbol{\mu}}$ is then utilized as the final denoised image and $\Bar{\boldsymbol{\sigma}}^2$ represents the uncertainty of $\Bar{\boldsymbol{\mu}}$.

\subsection{Noise-aware prior assignment} \label{Noise-aware_prior_assignment}

\begin{algorithm}[t]
\small
\caption{Algorithm of the Proposed ScoreDVI}\label{alg}
\begin{algorithmic}[1]
\Statex \textbf{Input:} GMM number $K$, MC samples $M$, MMSE Gaussian denoisers $\mathcal{G}_{1,\cdots K}$, $X$-net, $\Phi$-net, $Z$-net, $\Omega$-net, total iterations $T$, noisy image $\boldsymbol{y}$, $t=1$
\Statex \textbf{Output:} denoised image and its variance
\While{$t \leq T$}
\State Compute $\Theta$ from $X$-net, $\Phi$-net, $Z$-net and $\Omega$-net
\For{$k \leq K$}  
\State Compute the derivatives of $\text{E}_{q(x_{ik})}  \log p(x_{ik})$ in $\mathcal{L}_2$ to $\boldsymbol{\mu}_k(t)$ and $ \boldsymbol{\sigma}^2_k(t)$ based on Eq. \eqref{score_KL}, Eq. \eqref{score_KL2} and $\mathcal{G}_k$
\State Compute $\mathcal{L}_1$, $\mathcal{L}_3$, $\mathcal{L}_4$, $\mathcal{L}_5$ and the first item in $\mathcal{L}_2$
\State Compute the derivatives of $\Theta$ with regard to weights of four CNNs \Comment{This is done in autograd}
\State Update model weights with the optimizer
\EndFor
\EndWhile
\State \Return 
$\Bar{\boldsymbol{\mu}}(T)$ and $\Bar{\boldsymbol{\sigma}}^2(T)$ based on Eq. \eqref{posterior_ensemble}
\end{algorithmic}
\end{algorithm}

Observe that in the loss function $\mathcal{L}$ derived from standard VI, the image prior $p(\boldsymbol{x})$ and the likelihood are given equal weight. In practice, however, we expect the weighting between these two terms to depend on the specific input $\boldsymbol{y}$. For images with less noise, we would like the reconstruction results to be more biased toward the actual observations. On the other hand, for noisier images, we expect the restorations to be more dependent on the prior information. Therefore, we propose to adaptively adjust the weighting between the image priors and likelihoods based on the level of noise present in the input image. Specifically, we scale $\mathcal{L}_2$ to 
\begin{equation}
\mathcal{L}_2=\lambda(\boldsymbol{y})\text{E}_{q(Z)}\left(\text{KL}(q(X|Z||p(X|Z)\right)
\end{equation}
where $\lambda(\boldsymbol{y})$ is a unary weight function depending on $\boldsymbol{y}$.

To roughly estimate the noise level of $\boldsymbol{y}$, we use the noise estimation method presented in \cite{possion_gaussian_estimation} for generalized signal-dependent noise. Since real-world noise is spatially correlated, we first split $\boldsymbol{y}$ into down-scaled sub-images using a PD operation with a stride of 4. We then estimate the noise variance of local patches in these sub-images, following \cite{possion_gaussian_estimation}, and retain the patches with weak texture. The average noise standard deviation $\delta$ is then calculated from the remaining patches, providing a rough indication of the noise level presented in $\boldsymbol{y}$. 

Based on $\delta$, we make a simple schedule of $\lambda(\boldsymbol{y})$ to assign different beliefs of image priors to $\mathcal{L}$ in Eq. \eqref{ELBO}, that is
\begin{equation}\label{KL_reg}
    \lambda(\boldsymbol{y}) = \left \{\begin{array}{ll}
         \frac{1}{\gamma} &\text{ if } \delta < l_1\\
         1 &\text{ else if } l_1 \leq \delta < l_2 \\
         \gamma &\text{ else if } \delta \geq l_2
    \end{array} \right.
\end{equation}

\section{Experiments}

\begingroup
\renewcommand{\arraystretch}{1.1} 
\begin{table*}[h]
\caption{Quantitative comparisons (PSNR(dB)/SSIM) of our ScoreDVI and other real-world denoising methods including single image-based methods and dataset-based methods on
SIDD, FMDD, PolyU, and CC datasets. The best and second-best PSNR/SSIM results are marked in \textbf{bold} and \underline{underlined} in each denoising category. 
}
\label{comparison_single_image}
\small
\centering
\begin{tabular}{c|lccccc}
\hline
                 Category & Method &  SIDD validation&  SIDD benchmark &  FMDD & PolyU & CC \\ \hline
\multirow{7}{*}{\shortstack{Single image}} & BM3D \cite{BM3D} & 25.65/0.475 & 25.65/0.685 & 30.06/0.771  & 37.40/0.953  & 35.19/0.858 \\
              & DIP \cite{DIP} & 32.11/0.740 & -- & 32.90/0.854 & 37.17/0.912 & 35.61/0.912  \\
              & Self2Self \cite{Self2self} & 29.46/0.595 & 29.51/0.651 & 30.76/0.695 & \textbf{38.33/0.962} & \textbf{37.44/0.948} \\
              & PD-denoising \cite{PD} & \underline{33.97/0.820} & \underline{33.61/0.894} & \underline{33.01/0.856} & 37.04/0.940 & 35.85/0.923  \\
              & NN+denoiser \cite{zheng2020unsupervised} & -- & 33.18/0.895 & 32.21/0.831 & 37.66/0.956 & 36.52/0.943 \\
              & APBSN-single \cite{APBSN} & 30.90/0.818 & 30.71/0.869 &  28.43/0.804 & 29.61/0.897 & 27.72/0.891 \\
              & \textbf{ScoreDVI (Ours)} & \textbf{34.75/0.856} & \textbf{34.60/0.920} & \textbf{33.10/0.865} & \underline{37.77/0.959} & \underline{37.09/0.945} \\ \hline
\multirow{4}{*}{\shortstack{Noisy or Unpaired  \\images}} & APBSN \cite{APBSN} & -- & \textbf{36.91/0.931} & \underline{31.99/0.836} & \textbf{37.03}/\underline{0.951} & \underline{34.88/0.925} \\
                  & CVF-SID \cite{CVF-SID}& \underline{34.81}/\textbf{0.944} & 34.71/0.917 & \textbf{32.73/0.843} & 35.86/0.937 & 33.29/0.913 \\
                & LUD-VAE \cite{LUD_VAE} & \textbf{34.91}/\underline{0.892} & \underline{34.82/0.926} & -- & \underline{36.99}/\textbf{0.955} & \textbf{35.48/0.941} \\
                  \hline
\end{tabular}
\end{table*}
\endgroup

In this section, we first introduce the experimental settings. Then we present qualitative and quantitative results of our method as well as comparisons with other methods. The ablation study is carried out in the last.

\subsection{Experimental settings}

\noindent \textbf{Dataset}. We evaluate our ScoreDVI on four widely-used real-world noise datasets, namely SIDD \cite{SIDD}, PolyU \cite{polyu}, CC \cite{CC}, and FMDD \cite{FMDD}.
The SIDD dataset contains natural sRGB images from smartphones, and we evaluate our method using the SIDD validation and benchmark datasets, each of which consists of 1280 patches with size $3 \times 256 \times 256$. The PolyU and CC consist of 100 and 15 natural images, respectively, taken from diverse commercial camera brands. Each image in these datasets has a size of $3 \times 512 \times 512$. FMDD contains fluorescence microscopy images, and we evaluate our method on the mixed test set with raw images, which have an image size of $512 \times 512$.

\noindent \textbf{Implementation of MMSE Non-$i.i.d$ Gaussian denoisers}. We obtain MMSE Non-$i.i.d$ Gaussian denoisers using the bias-free (BF) \cite{BFCNN} version of DnCNN \cite{DNCNN} as the network architecture. We train multiple blind BF-DnCNNs over different clean datasets, including CBSD300 \cite{CBSD500}, DIV2K \cite{timofte2017DIV2k}, ImageNet \cite{deng2009imagenet} validation dataset, and Waterloo Exploration Database \cite{ma2016waterloo}, to implicitly learn diverse score priors for sRGB denoising. For each dataset, we add AWGN with noise standard deviation of $[0,100]$ to clean patches to construct noisy/clean pairs. The BF-DnCNN is then optimized with the MMSE objective using these paired data. We train BF-DnCNNs on corresponding grayscale images to capture scores for microscopy image denoising.

\noindent \textbf{Implementation of variational posteriors}. We construct four CNNs to represent the parameters $\Theta$ in Eq. \eqref{variational_posterior}. For the $X$-net, we use the Unet \cite{unet} architecture with skip connections, which outputs two maps, each with a shape of $\left(K,C,H,W\right)$, to denote $\boldsymbol{\mu}_{1,\cdots K}$ and $\boldsymbol{\sigma}^2_{1,\cdots K}$, respectively. For $\boldsymbol{\hat{\alpha}}_{1,\cdots K}$ and $\boldsymbol{\hat{\beta}}_{1,\cdots K}$, we build upon the DnCNN with 5 convolution layers as done in VDN \cite{VDN} and have the $\Phi$-net output two maps with the same shape as $\boldsymbol{\mu}_k$ and $\boldsymbol{\sigma}^2_k$. Regarding $Z$-net and $\Omega$-net, we use the same architectures as $\Phi$-net to output one map each to represent $\boldsymbol{\hat{d}}_{1,\cdots K}$ and $\boldsymbol{\omega}_{1,\cdots K}$.
Details of these networks can be found in Supplementary Material S2.

\noindent \textbf{Selection of hyperprior parameters}. 
We set $\alpha=1,\beta=0.02$ for the SIDD and FMDD datasets, which are known to contain highly noisy images. For CC and PolyU datasets, which have medium and low noise levels respectively, we set $\alpha=1,\beta=0.01$, and $\alpha=1,\beta=0.005$, respectively. We set $\boldsymbol{d}$ as an all-one vector in Eq. \eqref{GMM1}, to assign the same prior weight to each component.

\noindent \textbf{Optimization details}. During optimization, all
losses in $\mathcal{L}$ except $\mathcal{L}_2$ can be expressed as scalar functions in Pytorch \cite{paszke2019pytorch}. As for $\mathcal{L}_2$, we implement its gradient backpropagation by using a customized \textit{autograd} unit inherited from \textit{torch.autograd.Function}. Its backward function computes and returns the gradients based on Eqs. \eqref{score_KL} and \eqref{score_KL2}. Incorporating this unit into the total loss $\mathcal{L}$ enables end-to-end training of ScoreDVI in PyTorch. We use the ADAM optimizer with a fixed learning rate of $10^{-3}$ for all four networks. In Eq. \eqref{KL_reg}, we set $l_1=10$, $l_2=25$, and $\gamma=2$. We choose $K=3$, $M=5$, and perform the optimization with a total of 400  iterations for each noisy input. We evaluate the denoising quality using peak signal-to-noise ratio (PSNR) and structural similarity (SSIM) metrics. All denoising experiments are conducted on an Nvidia 2080 GPU. 

\subsection{Evaluation on real-world noise}

\begin{figure*}[!htb]
    \centering
    \includegraphics[width=0.93\textwidth]{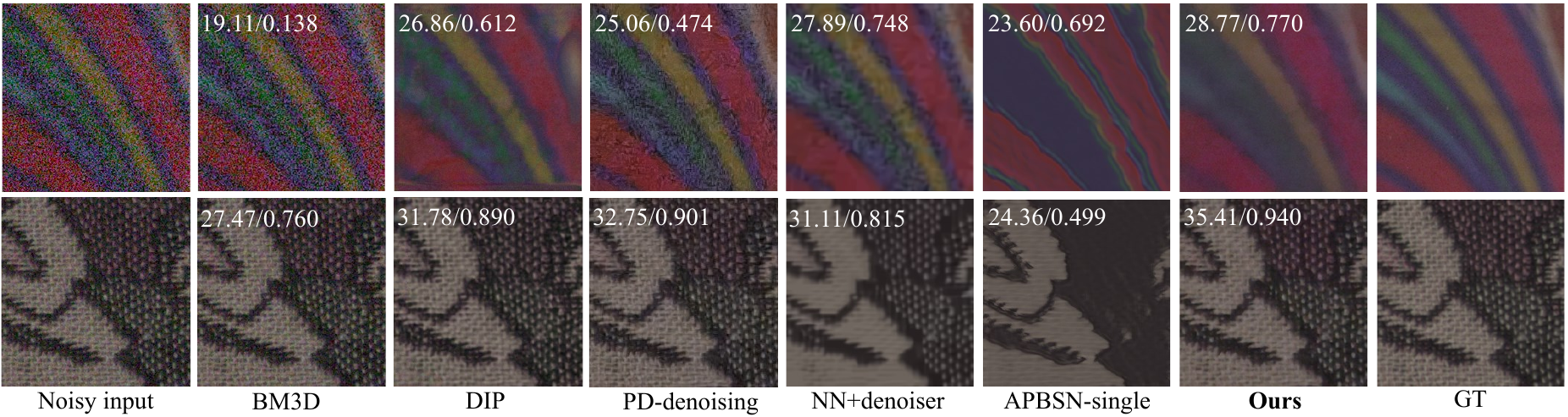}
    \caption{Visual comparison of our method against other single image-based denoising methods in SIDD validation dataset. The PSNR/SSIM results are listed within the images.}
    \label{fig:visual_comparison_SIDD_validation1}
\end{figure*}

\begin{figure*}[!htb]
    \centering
    \includegraphics[width=0.93\textwidth]{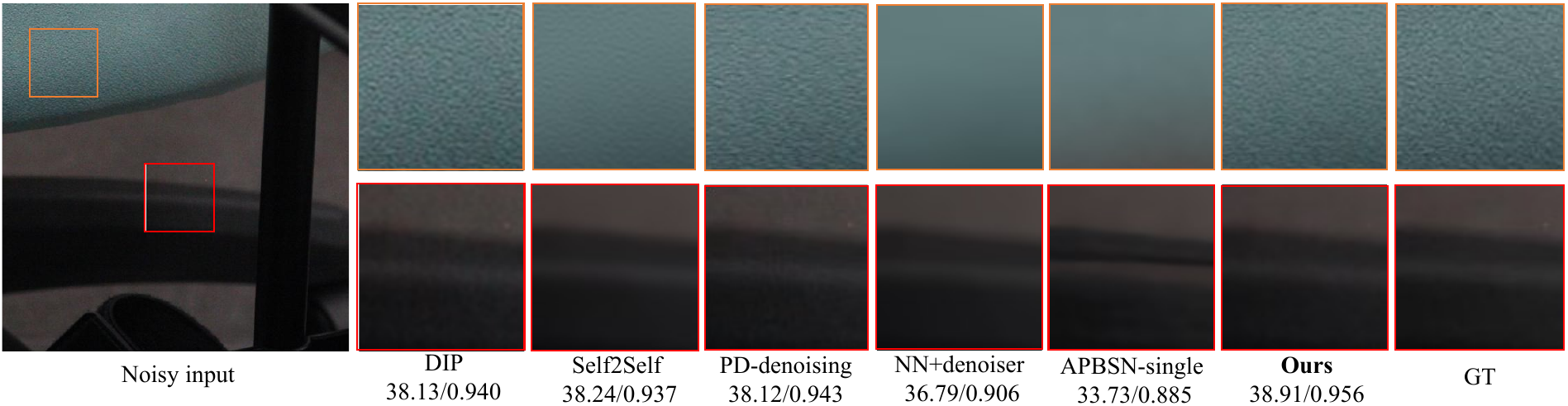}
    \caption{Visual comparison of our method against other single image-based denoising methods in PolyU dataset.}
    \label{fig:visual_comparison_polyu_validation1}
    
\end{figure*}

\begin{figure}
    \centering
\includegraphics[width=0.45\textwidth]{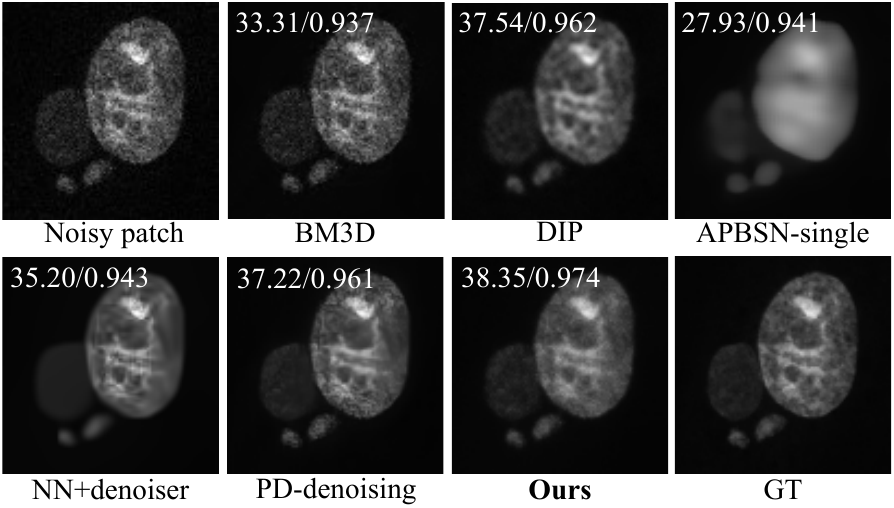}
    \caption{Visual comparison of single image-based denoising methods in FMDD. PSNR/SSIM are evaluated on the whole image. The noisy patch is from \text{Confocal\_BPAE\_B\_4}.}
\label{fig:visual_comparison_FMDD1}

\end{figure}

\noindent \textbf{Comparisons with single image-based methods}. We compare our method against several single image-based methods, including BM3D \cite{BM3D}, DIP \cite{DIP}, self2self \cite{Self2self}, PD-denoising \cite{PD}, NN+denoiser \cite{zheng2020unsupervised}, and APBSN-single \cite{APBSN}. 1) For DIP, we choose the same Unet architecture as our $X$-net. Due to its problem of overfitting, we choose the optimal denoising results with the reference to ground truth (GT) images. Hence, DIP in our experiments only serves as a reference. 
2) For APBSN-single, we adapt APBSN proposed in \cite{APBSN} to directly denoise a single image. The strides of PD in training and testing are 5 and 2, respectively. 3) For NN+denoiser \cite{zheng2020unsupervised}, we choose its best version, i.e. NN+BM3D for single image denoising. We re-evaluate the performance of NN+BM3D on FMDD, PolyU, and CC without the reference of GTs based on their source code. 4) For the remaining methods, we either use the authors’ code or directly adopt their published results if available. We summarize the quantitative comparisons in Table \ref{comparison_single_image}. Qualitative comparisons between different single image-based methods on SIDD, FMDD, and PolyU are shown in Figures \ref{fig:visual_comparison_SIDD_validation1}, \ref{fig:visual_comparison_FMDD1} and \ref{fig:visual_comparison_polyu_validation1}, respectively. More visual comparisons can be found in Supplementary Material S4.

Our ScoreDVI achieves the best quantitative and qualitative performance among other single image-based methods on the SIDD and FMDD, which contain noisier images. PD-denoising can effectively remove noise, but weedy artifacts are frequently observed in restored images, as indicated in Figures \ref{fig:visual_comparison_SIDD_validation1}, \ref{fig:visual_comparison_FMDD1}, and \ref{fig:visual_comparison_polyu_validation1}. Due to the large stride of PD used in training, APBSN-single severely destroys the details of original images and leaves clear color artifacts, as shown in Figures \ref{fig:visual_comparison_SIDD_validation1} and \ref{fig:visual_comparison_polyu_validation1}, or severe blur as shown in Figure \ref{fig:visual_comparison_FMDD1}. NN+denoiser introduces over-smoothing while removing noise. Self2Self performs badly on SIDD and FMDD, and cannot completely eliminate structured noise. In contrast, our method can both denoise noisy images and retain more details, thus yielding better results. Regarding the CC and PolyU datasets, Self2Self achieves the best PSNR with the cost of hundreds of thousands of training iterations. It also brings over-smoothed reconstructions, as shown in Figure \ref{fig:visual_comparison_polyu_validation1}. Instead, our ScoreDVI is quite efficient while also preserving fine details. Note that our method surpasses the optimal DIP among all datasets, as shown in Table \ref{comparison_single_image}.

\noindent \textbf{Comparisons with dataset-based methods.} Our ScoreDVI is also compared against several dataset-based unsupervised real-world denoising methods, including CVF-SID \cite{CVF-SID}, APBSN \cite{APBSN}, and LUD-VAE \cite{LUD_VAE}. APBSN and CVF-SID are directly trained on the test datasets, while LUD-VAE is trained on unpaired data where the clean data is from the SIDD small dataset and the noisy data are from the test datasets. We present the quantitative results in Table \ref{comparison_single_image}, and the visual comparisons can be found in Supplementary Material. Our method shows competitive results compared to dataset-based methods among these datasets. While the results of dataset-based methods outperform ours on SIDD validation and benchmark, their performance degrades with the decrease of available noisy images in datasets such as PolyU, CC, and FMDD. In this case, they can only utilize fewer available images to train their networks. In contrast, our method requires only a single noisy image and performs consistently over all datasets, making it more practical for real-world applications.

\noindent \textbf{Evaluating the existing score-based method for real-world denoising.} We evaluate one typical score-based method by combing techniques from \cite{score_sde} and \cite{score_MRI} for real-world denoising and present the result in Table \ref{score_based_methods_for_denoising}. We first employ the NCSN++ network and VESDE strategy proposed in \cite{score_sde} to train scores on ImageNet validation. To obtain posterior samples, we use the predictor-corrector sampler with 1000 steps, where the unconditional score in the original corrector is replaced with the posterior score for Gaussian denoising presented in \cite{score_MRI}. For each input, we obtain eight posterior samples and evaluate their means. By comparing the results in Table \ref{comparison_single_image} and \ref{score_based_methods_for_denoising}, we observe that our ScoreDVI outperforms this simple adoption of score-based methods for real-world structured noise by large margins.

\begin{table}
\small
\caption{Performance of the score-based method by combining  \cite{score_sde} and \cite{score_MRI} on SIDD validation and CC.}
\label{score_based_methods_for_denoising}
\centering
\begin{tabular}{ccc}
\hline
 Dataset & SIDD validation & CC  \\ \hline
 PSNR/SSIM & 32.30/0.769 & 36.18/0.936  \\ \hline
\end{tabular} 
\end{table}

\subsection{Ablation study}
Here, we conduct ablation studies on SIDD validation dataset to better evaluate the performance of our method.

\begin{figure}[h]
    \centering
\includegraphics[width=0.49\textwidth]{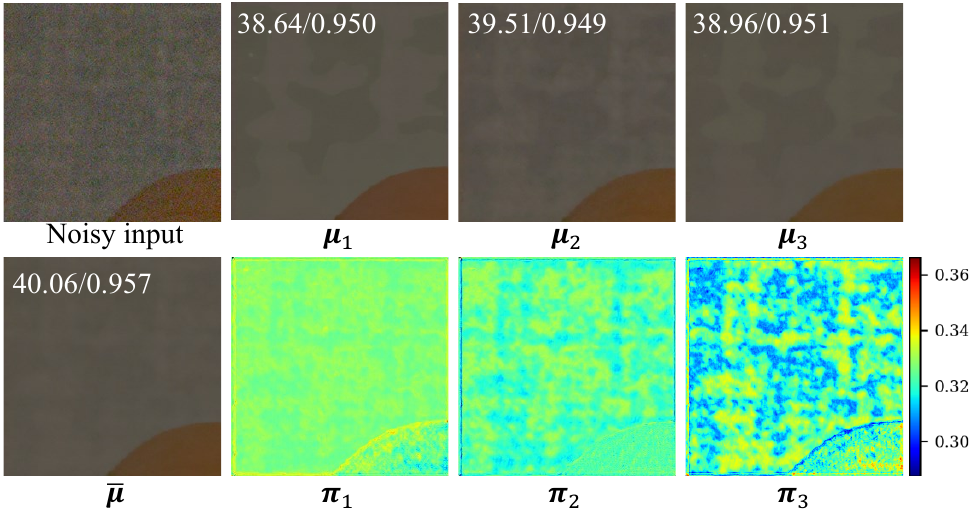}
    \caption{Visualization of different means $\boldsymbol{\mu}_k$ and $\boldsymbol{\pi}_k$ for a noisy sRGB image of SIDD validation dataset.}
\label{fig:ablation_GMM3}
\end{figure}

\noindent \textbf{Ablation on Gaussian numbers $K$}. Table \ref{ablation_K} presents the results of our method with different $K$ on the SIDD validation. As shown in Table \ref{ablation_K}, the denoising performance consistently improves as we increase $K$. Additionally, Figure \ref{fig:ablation_GMM3} shows that the ensemble of multiple means, $\boldsymbol{\mu}_k$, of the variational image posteriors, achieves better performance compared to using each single $\boldsymbol{\mu}_k$. Considering both performance and efficiency, we set $K=3$ in our method.  
\begin{table}[]
\caption{Effects of Gaussian mixture numbers $K$ and prior assignment coefficient $\gamma$ on SIDD validation dataset.}
\small
\centering
\subfloat[$\gamma=1$]{
\label{ablation_K}
\begin{tabular}{cccc}
\hline
$K$& PSNR/SSIM   \\ \hline
1 & 33.20/0.789   \\ 
2 & 33.68/0.808   \\ 
3 & 33.79/0.807   \\ 
4 & \textbf{33.87/0.815}   \\ \hline
\end{tabular} 
}
\quad\quad\quad
\subfloat[$K=3$]{
\label{ablation_gamma}
\begin{tabular}{cccc}
\hline
$\gamma$ & PSNR/SSIM   \\ \hline
1 & 33.79/0.807  \\  
$4/3$ & 34.55/0.852   \\ 
2 & \textbf{34.75/0.856}   \\
4 &  33.83/0.831   \\ \hline
\end{tabular} 
}
\end{table}

\begin{figure}[]
    \centering
\includegraphics[width=0.48\textwidth]{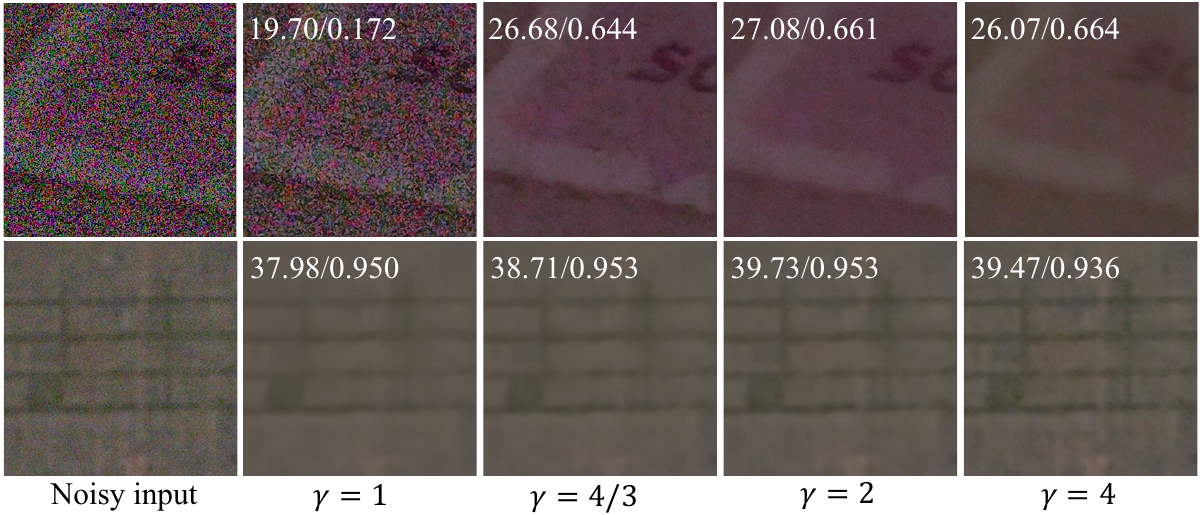}
    \caption{Visualization of our method under different $\gamma$ for noisier (first row) and less noisy (second row) images.}
    \label{fig:ablation_klreg}
\end{figure}

\noindent \textbf{Ablation on prior assignment coefficient $\gamma$}. Table \ref{ablation_gamma} presents the results of our method with different $\gamma$. From Table \ref{ablation_gamma} and Figure \ref{fig:ablation_klreg}, we have the following observations: when $\gamma=1$, it means no adaptive prior assignment, and the corresponding denoised images either contain much noise for the noisier image (the first row of Figure \ref{fig:ablation_klreg}) or are over-smooth for the less noisy image (the second row of Figure \ref{fig:ablation_klreg}); with the increase of $\gamma$, the noise is gradually removed, and details emerge; when $\gamma$ becomes too large, either image priors or actual observations dominate the optimization, and the denoised images show artifacts or become noisy. Therefore, we set $\gamma=2$ for both clean and sharp results.  

\noindent \textbf{Ablation on the modeling of parameters $\Theta$}. We evaluate the effect of using CNNs to model $\Theta$ on our method's performance. We compare our method and the one that directly uses optimizable $\Theta$ initialized from $rand$ Pytorch tensors to minimize $\mathcal{L}$. The results in the first and third row of Table \ref{ablation_others} demonstrate the effectiveness of deep modeling of $\Theta$.   

\begin{table}[t]
\small
\caption{Effects of deep modeling of $\Theta$ (A), image-wise fusion (B), and pixel-wise fusion (C) on SIDD validation.}
\label{ablation_others}
\centering
\begin{tabular}{cccc}
\hline
A & B & C & PSNR/SSIM \\ \hline
 &  & \checkmark &  31.76/0.767   \\
\checkmark & \checkmark  &  & 34.34/0.850   \\
 \checkmark &  & \checkmark & \textbf{34.75/0.856}    \\ \hline
\end{tabular} 
\end{table}

\noindent \textbf{Ablation on the ensemble strategy of image priors and posteriors}. 
We evaluate the effectiveness of the pixel-wise fusion in Eq.  \eqref{posterior_ensemble} by comparing it with an image-level fusion strategy that bases on the image-level mixing of GMM for the likelihood. The final denoised image is hence the image-wise fusion of variational image posteriors. Details of this strategy are presented in Supplementary Material S3. The second and third row in Table \ref{ablation_others} shows that the pixel-wise fusion achieves better results. 

\subsection{Efficiency comparisons and empirical convergence analysis}
We present the inference time, model parameters, and FLOPs of the compared deep learning-based methods in Table \ref{complexity_comparisons}. By comparing Table \ref{comparison_single_image} and Table \ref{complexity_comparisons}, our method achieves a balance between effectiveness and efficiency.
\begin{table}[h]
	\centering
    \small
	\caption{Efficiency comparisons of deep learning-based methods under the input size $256\times 256\times 3$}
	\begin{tabular}{cccc}
		\hline
		Method & Infer. time (s) & Params (M) & FLOPs (G)\\ \hline
		DIP  & 146.2 & 13.4 & 31.06 \\
		Self2Self  & 3546.5 & 1.0 & 9.55 \\
		PD-denoising & 0.36 & 0.7 & 46.94 \\
		NN+denoiser & 897.6& 13.4 & 31.06\\ 
		APBSN-single & 121.4 & 3.66 & 234.63 \\
		ScoreDVI & 81.2 & 13.5 & 37.87\\ \hline
	\end{tabular}
 \label{complexity_comparisons}
\end{table}

We ablate different iteration numbers and show the corresponding results (PSNR/SSIM) in Table \ref{Optimization_iterations}. As shown, increasing the iteration number results in the convergence of performance and does not show overfitting issues. We find that 400 iterations are sufficient to achieve a balance between performance and efficiency \textit{on average}, but it may be not the optimal number for each input.
\begin{table}[t]
	\centering
    \small
    \caption{Denoising performance of the proposed ScoreDVI with regard to different iteration numbers}
    \label{Optimization_iterations}
    \begin{tabular}{ccc}
        \hline
        Iters. & SIDD Val & CC \\ \hline
        300 & 34.46/0.847 & 36.91/0.943 \\
        400 & 34.75/0.856 & 37.09/0.945 \\ 
        500 & 34.78/0.861 & 37.13/0.946 \\
        600 & 34.75/0.861 & 37.12/0.946 \\
        \hline
    \end{tabular}
\end{table}

\section{Conclusion}
In this paper, we propose a novel unsupervised deep variational inference method that incorporates
score priors embedded in MMSE Non-$i.i.d$ denoisers and the Non-$i.i.d$ Gaussian mixture likelihood model for real-world single image denoising. A noise-aware priors assignment strategy is developed to further improve denoising performance. Our proposed method shows superior performance compared with other single image-based  real-world denoising methods on a variety of benchmark datasets. Moreover,
our approach only requires a single noisy image and is more generalizable than dataset-based un/self-supervised real-world denoising methods.

\section{Acknowledgment}
This work was supported in part by the National Natural Science Foundation of China (NNSFC), under Grant Nos. 61672253 and 62071197.

{\small
\bibliographystyle{ieee_fullname}
\bibliography{egbib}
}

\end{document}